\documentclass[acmtog,screen]{acmart}
\usepackage{multirow}

\AtBeginDocument{%
  \providecommand\BibTeX{{%
    \normalfont B\kern-0.5em{\scshape i\kern-0.25em b}\kern-0.8em\TeX}}}

\setcopyright{acmcopyright}
\copyrightyear{2023}
\acmYear{2023}
\acmDOI{XXXXXXX.XXXXXXX}

\acmConference[Conference acronym 'XX]{Make sure to enter the correct
  conference title from your rights confirmation emai}{June 03--05,
  2018}{Woodstock, NY}
\acmBooktitle{Woodstock '18: ACM Symposium on Neural Gaze Detection,
 June 03--05, 2018, Woodstock, NY} 
\acmPrice{15.00}
\acmISBN{978-1-4503-XXXX-X/18/06}

\begin{document}

\title{Dance Generation by Sound Symbolic Words}

\author{Miki Okamura}
\email{mikio_kamura@digitalnature.slis.tsukuba.ac.jp}
\orcid{0009-0001-9367-1078}
\affiliation{%
  \institution{University of Tsukuba}
  \streetaddress{}
  \city{Tsukuba}
  \state{Ibaraki}
  \country{Japan}
  \postcode{305-0821}
}

\author{Naruya Kondo}
\email{n-kondo@digitalnature.slis.tsukuba.ac.jp}
\affiliation{%
  \institution{University of Tsukuba}
  \streetaddress{}
  \city{Tsukuba}
  \state{Ibaraki}
  \country{Japan}
}

\author{Tatsuki Fushimi}
\affiliation{%
  \institution{University of Tsukuba}
  \streetaddress{}
  \city{Tsukuba}
  \country{Japan}}
\email{tfushimi@digitalnature.slis.tsukuba.ac.jp}

\author{Maki Sakamoto}
\affiliation{%
 \institution{The University of Electro-Communications}
 \streetaddress{}
 \city{Chofu}
 \state{Tokyo}
 \country{Japan}}

\author{Yoichi Ochiai}
\affiliation{%
  \institution{University of Tsukuba}
  \streetaddress{}
  \city{Tsukuba}
  \state{Ibaraki}
  \country{Japan}}


\begin{abstract}
This study introduces a novel approach to generate dance motions using onomatopoeia as input, with the aim of enhancing creativity and diversity in dance generation. Unlike text and music, onomatopoeia conveys rhythm and meaning through abstract word expressions without constraints on expression and without need for specialized knowledge. We adapt the AI Choreographer framework and employ the Sakamoto system, a feature extraction method for onomatopoeia focusing on phonemes and syllables. Additionally, we present a new dataset of 40 onomatopoeia-dance motion pairs collected through a user survey. Our results demonstrate that the proposed method enables more intuitive dance generation and can create dance motions using sound-symbolic words from a variety of languages, including those without onomatopoeia. This highlights the potential for diverse dance creation across different languages and cultures, accessible to a wider audience.
Qualitative samples from our model can be found at: \url{https://sites.google.com/view/onomatopoeia-dance/home/}.
\end{abstract}



\begin{teaserfigure}
  \includegraphics[width=\textwidth]{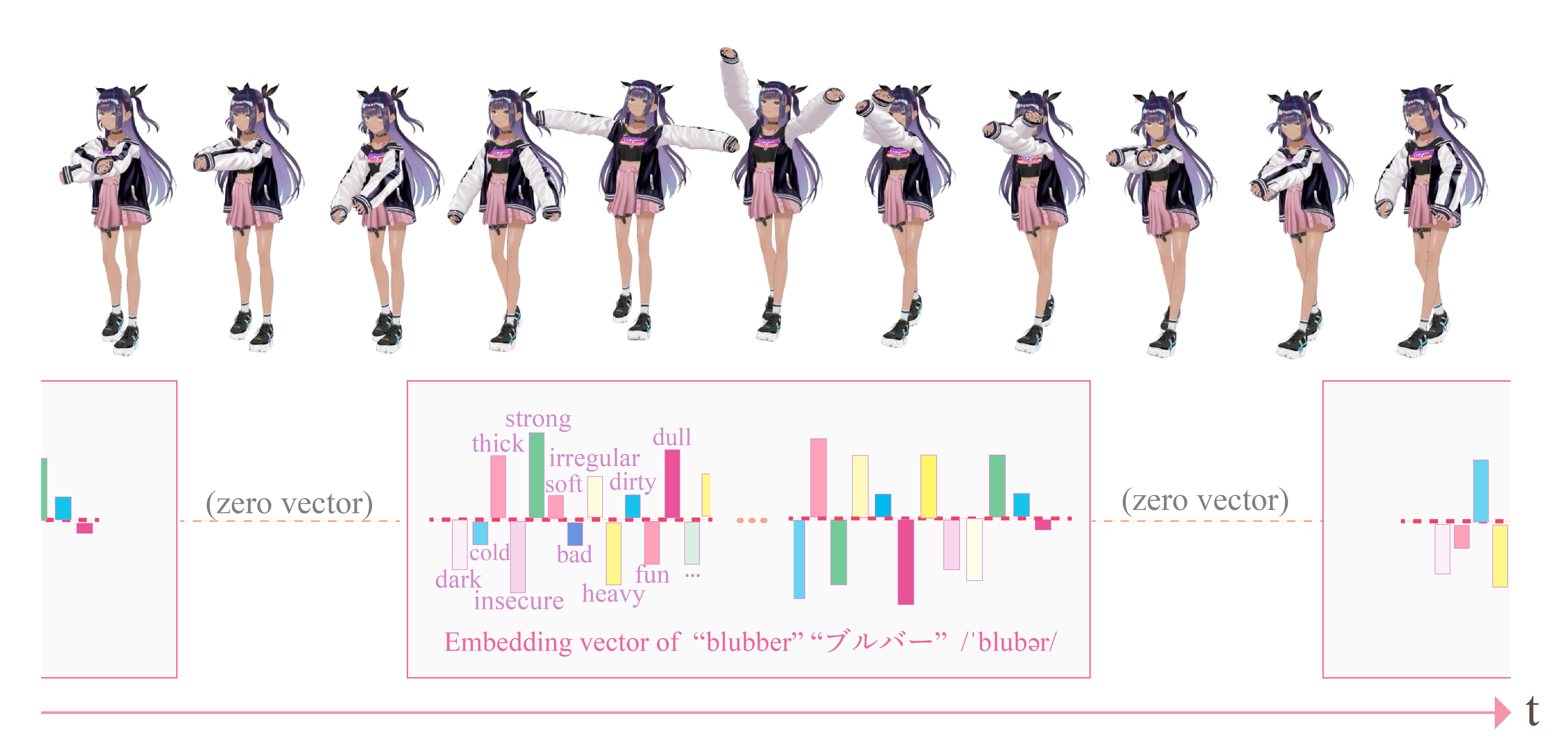}
  \caption{The dance motion generated by our model: By providing a creative onomatopoeia (in this case, "Bulber") as input, we can generate a dance that suits it. The input onomatopoeia is embedded in terms of the intensity of 43 adjective expressions by the Sakamoto system~\cite{doizaki2016automatic} and is input into a deep learning-based dance generation model called FACT~\cite{li2021ai}, which generates the dance. A zero vector is input during times when no onomatopoeia is provided.}
  \label{fig:teaser}
\end{teaserfigure}



\maketitle

\section{Introduction}
Dance is a universal form of expression that goes beyond the limits of language, culture, and geography. As an essential aspect of human history, dance has continually evolved with advancements in technology. One such development is the incorporation of computer graphics and animation in the automatic generation of dance. This combination of technology and creativity has the potential to offer new perspectives, support the creation of dance, and reduce effort. Furthermore, it contributes to developing diverse movements for interactive characters in video games and virtual reality applications.
With the progress of technology, machine learning techniques have been introduced to the field of dance generation. Traditional text-based dance generation methods can only handle expressions that exist in language, making it challenging to incorporate new sounds and sensations not represented in existing language. Consequently, these methods face difficulties in generating unique dance motions that go beyond the constraints of language and culture, limiting creativity and diversity. On the other hand, music-based dance generation methods, while rich in rhythm, may not effectively convey meaning. Moreover, these approaches are not easily accessible to users who cannot create music themselves.

We propose the utilization of onomatopoeia as a effective and innovative input source for a machine learning model aimed at generating dance motion. Onomatopoeia possesses the unique ability to convey both rhythm and meaning through abstract word sense expressions, making it an ideal candidate for this purpose. Its versatile nature allows it to be employed in various contexts, such as describing the texture of a meal, the touch of a cat, the speed of a car, or the ambiance of a comic strip, emphasizing its broad applicability and potential for enhancing the dance generation process.
Building upon the framework of an existing study named AI Choreographer~\cite{li2021ai}, we adapt this method to incorporate onomatopoeia as input for generating dance motion. In contrast to text and music inputs, we utilize the Sakamoto system~\cite{doizaki2016automatic} for extracting features from onomatopoeia. The system is a deterministic feature extraction method focusing on phonemes and syllables, representing onomatopoeia through 43 numerical values corresponding to the strength of adjective expressions. This system can be used with any language containing phonemes, as it is not influenced by linguistic meaning. Combining the Sakamoto system with a deep generative model is a novel approach, as there are few examples of such implementation in the literature. We collected 44 pairs of onomatopoeia and dance motions through a user survey and trained the model on this dataset. The results of our study indicate that the proposed model successfully generates dance motions that correspond with the input onomatopoeia. Furthermore, we demonstrate that the model can generate dances using sound-symbolic words from various languages, including those without onomatopoeia, allowing for more intuitive dance generation.
However, it must be acknowledged that the quality of the generated motions is not on par with the latest state-of-the-art dance generation methods.
Future research and development can focus on model improvements, dataset expansion, and the development of interactive applications for onomatopoeia-based dance motion generation.

This paper's contributions are as follows:
\begin{itemize}
    \item We introduce a novel approach using onomatopoeia as input for dance motion generation.
    \item  We adapt the AI Choreographer framework and employ the Sakamoto system for feature extraction from onomatopoeia.
    \item We present a new dataset of 44 onomatopoeia-dance motion pairs collected from a user survey.
    \item Our study demonstrates the successful generation of dance motions using sound-symbolic words from various languages, including those without onomatopoeia.
    \item The results indicate potential for intuitive and diverse dance creation across different languages and cultures.
\end{itemize}

\def \rela {EDGE (2022)~\cite{tseng2022edge}}
\def \relb {AI Choreographer (2021)~\cite{li2021ai}}
\def \relc {Yamagata et al. (2019)~\cite{yamagata2019image}}
\def \reld {Sakamoto et al. (2017)~\cite{sakamoto2017exploring}}
\def \rele {Doizaki et al. (2016)~\cite{doizaki2016constructing}}
\def \relf {Crnkovic-Friis et al. (2016)~\cite{crnkovic2016generative}}
\begin{table*}[t]

\centering
\begin{tabular}{lcc}
\toprule
& input & output or function \\
\midrule

\relf       & Text and Style & {\bf Dance} \\
\rela       & Music & {\bf Dance} \\
\relb       & Music & {\bf Dance} \\
{\bf Ours}  & {\bf Sound Symbolic Words} & {\bf Dance} \\
\relc       & {\bf Sound Symbolic Words} & Image retrieval \\
\reld       & {\bf Sound Symbolic Words} & Material retrieval \\
\rele       & {\bf Sound Symbolic Words} & Medical conditions \\

\bottomrule
\end{tabular}
\caption{Overview of related works. We have summarized studies that are closely related to our research, involving dance as the output and sound symbolic words as the input. Currently, we are exploring new combinations within these fields.}
\label{tab:relworks}
\end{table*}
\section{Related Work}

\subsection{Human Motion Generation}
The problem of human motion generation has been extensively studied in the fields of computer vision, graphics, and robotics. Early approaches focused on employing statistical models, such as kernel-based probability distribution \cite{pullen2000animating, bowden2000learning, galata2001learning, brand2000style}, and motion matching techniques \cite{holden2020learned} to synthesize motion. However, these methods tended to abstract away motion details and were primarily restricted to simple domains like locomotion. The advent of deep learning has led to a surge of interest in exploring the applicability of neural networks for generating 3D motion by training on large-scale motion capture datasets. Various network architectures have been investigated, including CNNs \cite{holden2015learning, holden2016deep}, GANs \cite{hernandez2019human}, RBMs \cite{taylor2009factored}, RNNs \cite{fragkiadaki2015recurrent, aksan2019structured, jain2016structural, ghosh2017learning, chiu2019action, du2019bio, wang2019imitation, butepage2017deep, villegas2018neural}, and Transformers \cite{aksan2020attention, bhattacharya2021text2gestures, ling2020character, won2022physics}. While these approaches have demonstrated the capability to generate more diverse and realistic human motions than earlier methods, they still face challenges in capturing the physical laws governing human movement. MDM~\cite{tevet2022human} is a motion generation method using diffusion models \cite{ho2020denoising}, which has become capable of producing quite human-like movements. These motion generation models, while skilled at producing common human actions, are not well-suited for generating expressive and creative motions. Generating human motion conditioned on various inputs has also become an active area of research. Studies have explored conditioning on joystick control \cite{ling2020character}, class labels \cite{guo2020action2motion, petrovich2021action}, text descriptions \cite{petrovich2022temos, zhang2022motiondiffuse}, and seed motions \cite{duan2022unified, holden2016deep, rempe2021humor, yin2022dance}. These methods enable motion generation under various control conditions. However, unlike existing researches, our study focuses on generating motions conditioned by sound-symbolic words, which is the first attempt of its kind.

\subsection{Dance Generation}
The challenging task of generating dance motions that are stylistically faithful to input music has attracted the attention of many researchers. Early approaches in this domain followed a motion retrieval paradigm \cite{fan2011example, lee2013music, ofli2011learn2dance} or used optimization-based approaches \cite{tendulkar2020feel} to generate 2D pose skeletons from conditioning audio. However, these methods often resulted in unrealistic choreographies that lacked the complexity of human dances. More recent works have shifted towards synthesizing motion from scratch by training on large datasets. Various modeling approaches have been proposed like LSTMs \cite{alemi2017groovenet, tang2018dance, yalta2019weakly, zhuang2020towards, kao2020temporally}, GANs \cite{lee2019dancing, sun2020deepdance, ginosar2019learning}, convolutional sequence-to-sequence models \cite{ahn2020generative, ye2020choreonet}, normalizing flow \cite{valle2021transflower}, transformer \cite{huang2020dance, huang2022genre, kim2022brand, li2022danceformer, li2021ai, siyao2022bailando} and diffusion models \cite{tseng2022edge}.
Among these methods, recent developments like AI Choreographer~\cite{li2021ai} and EDGE~\cite{tseng2022edge} have started to generate diverse and human-like dances based on music inputs. However, because these methods require music for creating dance, they are not suitable for those who simply want to generate dances without providing music. Moreover, it has been challenging to casually edit input data or output dances, preventing anyone from intuitively generating dances with ease. While EDGE~\cite{tseng2022edge} has implemented editing features to simplify dance generation, enabling partial masking and generating under certain constraints, it can be difficult to master without some dance experience. On the other hand, we propose using sound-symbolic words as the input, allowing anyone to easily generate dances by merely entering intuitive words.

\subsection{Leveraging Onomatopoeia Representation}

In recent years, there has been a growing interest in research on utilizing onomatopoeic representations for applications such as search and prediction. Onomatopoeic words can represent abstract concepts such as the texture and feeling of an object or the rhythm of a sound, which are unique features not present in other forms of expression like language or music. Consequently, onomatopoeia can be considered an ideal medium for handling information in a more sensory-oriented manner.

Several studies have explored the use of onomatopoeia for data visualization and search, such as \cite{yamagata2019image, doizaki2013intuitive, sakamoto2017exploring, doizaki2015possibility, doizaki2016constructing, sakamoto2021automatic}. For instance, \cite{yamagata2019image} proposes an image retrieval system that uses sound-symbolic words to represent surface texture, while \cite{doizaki2013intuitive} develops a color design support system that quantifies images using onomatopoeia to recommend suitable colors for intuitive and ambiguous design images. \cite{sakamoto2017exploring} introduces a method for selecting material samples using sensory vocabulary, effectively exploring the tactile perceptual space. \cite{doizaki2015possibility} presents a product search system that uses onomatopoeia to express rich product textures and rank related images accordingly. \cite{doizaki2016constructing} constructs a system that proposes metaphors corresponding to the onomatopoeia expressing medical conditions, enhancing the quality of medical interviews. \cite{sakamoto2021automatic} develops a system for estimating multidimensional personality traits from a single sound-symbolic word, contributing to the prediction of personality evaluations and the efficiency of communication. Furthermore, \cite{yamagata2021computer} proposes a deep convolutional neural network (DCNN)-based computer vision method that generates onomatopoeic expressions of material textures from input images, demonstrating human-like texture representation through onomatopoeia. In addition, \cite{nozaki2021sound} introduces a game system that incorporates onomatopoeia as a game controller, comparing and evaluating the user experience with traditional controllers and confirming an improved sense of enjoyment and novelty. While there are several search and generation methods using onomatopoeia, to the best of our knowledge, there have been no attempts at content creation utilizing onomatopoeia. This study presents the first endeavor in onomatopoeia-based generation, aiming to extract the great potential of onomatopoeia as a form of expression.

\section{Pilot Study and Data Collection} \label{userExp}

\subsection{Questionnaire survey}

To confirm the relationship between sound symbols a.k.a onomatopoeia and dance choreography and the possibility of generating dance movements with onomatopoeia, we conducted a questionnaire on dance and onomatopoeia among experienced dancers before and after the data collection experiment.In preliminary survey, participants answered the following questions: "Do you use onomatopoeia during dance practice? The options for whether or not they use onomatopoeia were "use frequently," "use often," "occasional use," and "I don't use it". After the annotation experiment, participants were asked if it was easy for them to annotate the onomatopoeia to the dance choreography. The options were "fairly easy," "easy," "somewhat easy," "normal," "somewhat difficult," "difficult," and "fairly difficult. Finally, respondents answered free-form questions about their impressions and findings of the annotations.

\subsubsection*{Questionnaire results}

\begin{figure*}[htbp]
    \centering
    \setlength{\tabcolsep}{12pt} 
    \begin{tabular}{cc}
    \centering
    \begin{minipage}[b]{0.45\linewidth}
        \centering
        \includegraphics[width=0.66\linewidth]{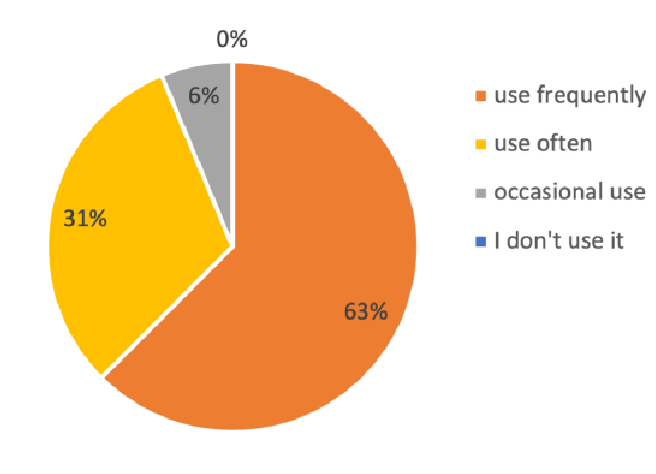}
        \caption[Questionnaire on the frequency of onomatopoeia use]{A questionnaire that asked respondents if they use onomatopoeia during dance practice in the pre-survey. Many respondents answered that they use onomatopoeia "often" or "often. Because one person did not respond, the result is the result of 15 respondents.}
        \label{fig:onomatope-use}
    \end{minipage} &
    \centering
    \begin{minipage}[b]{0.45\linewidth}
        \centering
        \includegraphics[width=0.66\linewidth]{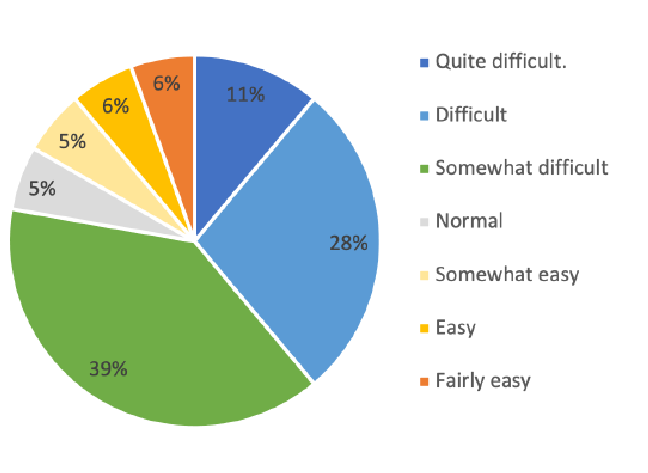}
        \caption[Questionnaire on the task of recalling onomatopoeia]{Post questionnaire asked whether it was easy to imagine onomatopoeia for the dance choreography. Many responded "somewhat difficult" or "difficult".}
        \label{fig:onomatope-diffi}
    \end{minipage}
    \end{tabular}
\end{figure*}

In the questionnaire asking whether they use onomatopoeia during dance practice, 62.5\% answered "often" and 32.3\% answered "yes," and in the questionnaire asking about the degree of difficulty of annotation, 47.1\% answered "somewhat difficult" and 23.5\% answered "difficult. The results indicated that although they may use onomatopoeia spontaneously during dance practice, it is difficult to imagine onomatopoeia for each movement

\subsection{Data collection}
In order to construct a dataset of dance and onomatopoeia to be used for training data and to explore the relationship between dance and onomatopoeia, we conducted a subject experiment in which we collected data on what kind of onomatopoeia was recalled in response to what kind of choreography and asked participants to describe their impressions and findings.
Japanese speakers aged 18-69 with dance experience were asked to annotate the onomatopoeia they recall in response to dance videos.

Data on onomatopoeia for dance was collected by selecting silent dance videos mainly in the genre of the subjects' experience and adding onomatopoeia to the dance choreography as subtitles using Youtube Studio's subtitle creation function. Dance videos from the front viewpoint of Advanced Dance in the AIST Dance Video Database\cite{aist-dance-db} were used to generate a variety of dance motions.
In the annotations, the following instructions were given: ``Annotate the onomatopoeia that came to you intuitively rather than carefully, and divide the dance movements into as many smaller pieces as possible to make the onomatopoeia.
Separate onomatopoeia and dance movements as finely as possible.
Annotate the choreography with onomatopoeia that comes to mind freely. You are free to create your own onomatopoeia, not limited to existing common vocabulary.''

During the one-hour experiment, the participants were asked to annotate one to five dance videos, depending on their progress.
This subject experiment yielded the annotation data for a total of 44 dance videos in 10 dance genres.

\section{System}

\begin{figure*}[htbp] 
\centering
\includegraphics[width=\linewidth]{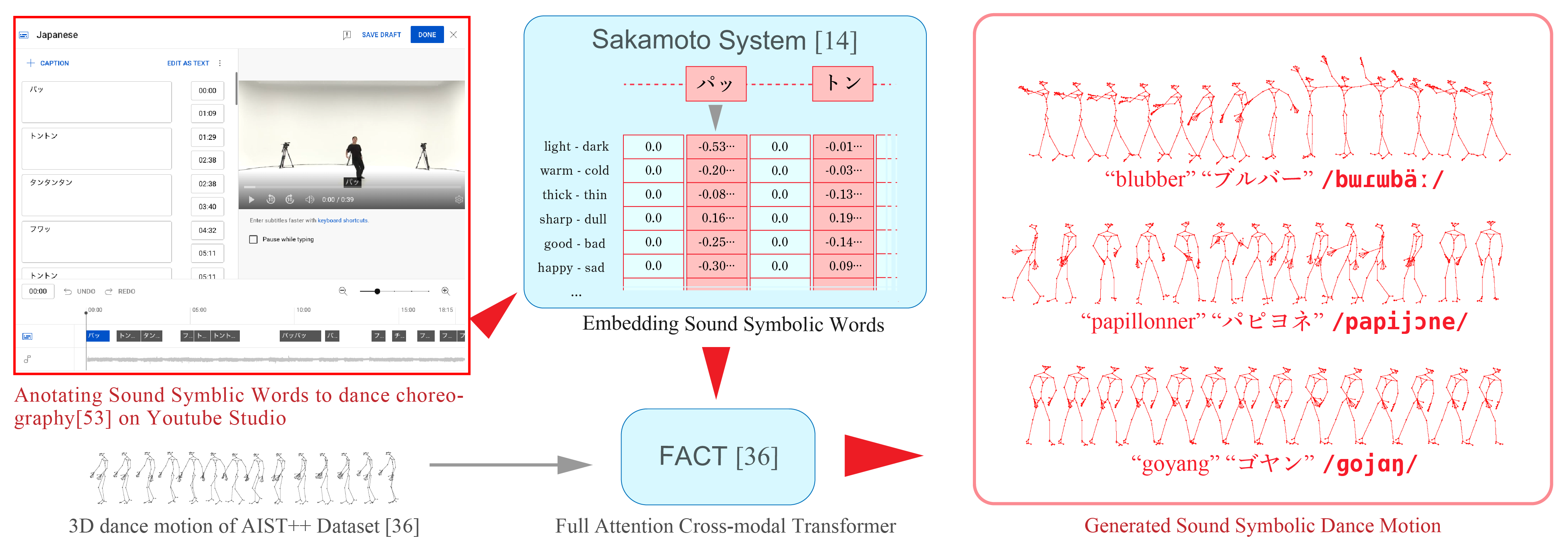}
\caption[Learning model]{\emph{Learning model} The onomatopoeia for dances collected in the subject experiments are quantified using Sakamoto System, which quantifies the fine impressions of each onomatopoeia using a scale of 43-dimensional adjective pair scale, and the data set is prepared as one onomatopoeia sequence per one dance, which became input dataset with dance motions\cite{li2021ai}. the output is the dance motion conditioned on the onomatopoeia. }
\label{fig:input-output}
\end{figure*}

In this study, we collected annotation data for AIST Dance Database\cite{aist-dance-db}, using the method described in \autoref{userExp}. 
The onomatopoeia collected in the subject experiments was quantified using sound symbol estimation system which we call `Sakamoto System'.
The quantified onomatopoeia were arranged in a time series and converted into a series vector to be used during training.
AIST++, which is three-dimensional motion data corresponding to the AIST Dance Database, and the series vectors were trained on the FACT model, and the motion was output. The following is a detailed description of the methods and techniques used to create the datasets treated as training data.

\subsection{Preparation and pre-processing of training data}
\subsubsection{Preparation}

In preprocessing, the collected onomatopoeia is converted into a 43-dimensional series vector corresponding to the dance motion and made into sequence data.
First, all the annotated onomatopoeia described in the caption file created using the subtitle editor of the video service Youtube in the subject experiment of the \ref{userExp} chapter are extracted, and the impression of the onomatopoeia is quantified by Sakamoto System. The system converted each onomatopoeia into a 43-dimensional numerical vector, and created an onomatopoeia quantification dictionary that describes the 43-dimensional numerical impression of each onomatopoeia.
Next, an array of 43 dimensions per index, each with 0 elements, is created, with the series length of the onomatopoeia series vector being the series length of the music corresponding to the dance motion in the previous research\cite{li2021ai} implementation.

Next, the onomatopoeia of the input is quantified from the numeric dictionary, and the time (in seconds) when each onomatopoeia is given as a subtitle is multiplied by fps=60 to obtain the number of frames, and the features of the corresponding index number of the series vector are replaced.
For example, if the onomatopoeia "kuru-kuru" is annotated between 2.0 and 4.0 seconds, the index numbers from 120 to 240 of the series vector are replaced with a 43-dimensional numeric vector representing the "kuru-kuru" impression.

The sample of the onomatopoeia series vector is the following array where \begin{math}d=43\end{math} is the 43-dimensional feature of the adjective pair calculated by Sakamoto System, \begin{math}t\end{math} is the number of frames corresponding to the motion video The number of frames is the number of frames corresponding to the motion video.

\begin{figure}[htbp] 
\centering
\includegraphics[width=0.7\linewidth]{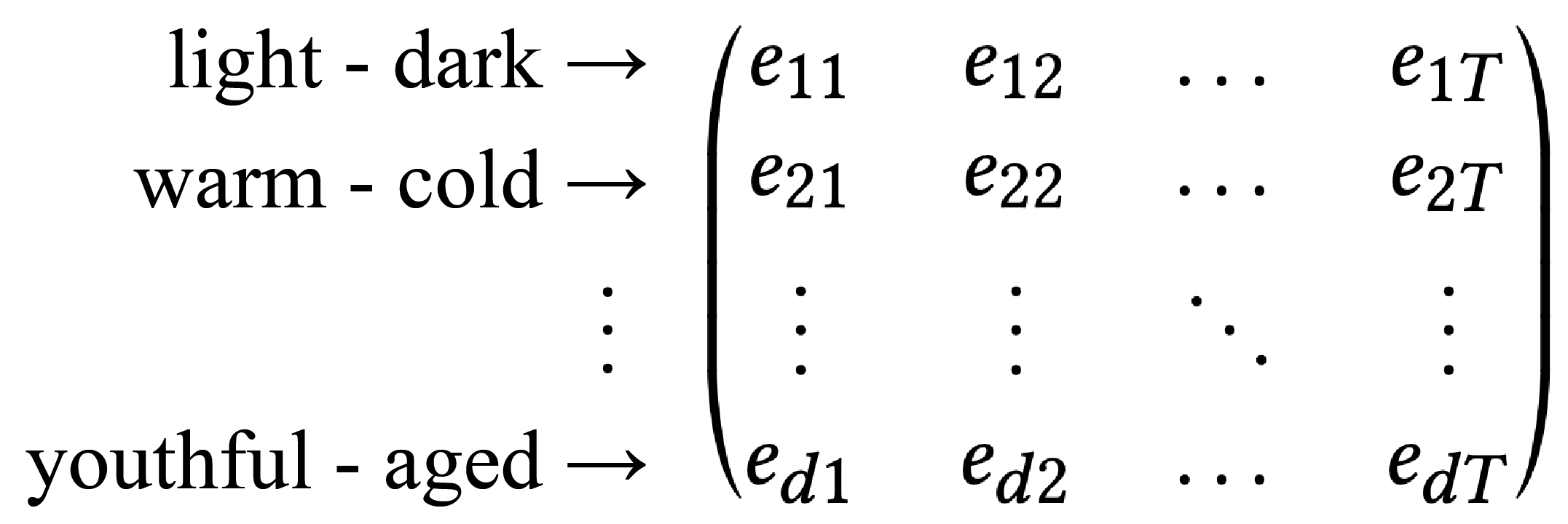}
\end{figure}

\subsubsection{Sakamoto System\cite{doizaki2016automatic}: Embedding sound symbolism} \label{quantication}

To quantify onomatopoeia, we use Sakamoto System for estimating fine impressions of onomatopoeia. Based on the phonetic symbolism of the Japanese language, the System quantifies the image of the input onomatopoeia using a scale of 43 adjective pairs related to mainly visual and tactile perception, for example light - dark, warm - cold, sharp - mild, and heavy - light.

The system is able to estimate the wealth of information conveyed by not only existing onomatopoeia but also newly created onomatopoeia, and to distinguish between a variety of highly similar onomatopoeia. For this reason, in the subject experiments in this study, participants were asked to freely respond to the onomatopoeia they imagined from the action.

Since the system used in this study does not quantify the impression of onomatopoeia in real time, the onomatopoeia collected in the subject experiment was listed and quantified in advance to create an onomatopoeia dictionary, and the impression was quantified through the onomatopoeia dictionary during preprocessing and generation at the time of evaluation. The onomatopoeic dictionary is then used for preprocessing and generation at the time of evaluation to create onomatopoeic series vectors.

\subsection{Learning and evaluation}
We used FACT, a model used in AI Choreographer, which generates dance motions from music. The input layer, which encodes music in 35 dimensions, was changed to encode onomatopoeia in 43 dimensions, and the collected data was trained with FACT.

The input of the model contains a seed motion sequence with 120 frames (2 seconds) and a music sequence with 240 frames (4 seconds), where the two sequences are aligned on the first frame. The output of the model is the future motion sequence with N = 20 frames supervised by L2 loss. During inference FACT continually generate future motions in a autoregressive manner at 60 FPS, where only the first predicted motion is kept in every step.

For the learning and evaluation phase, we employed the cross-modal learning of sound symbols (onomatopoeia) and dance motions using FACT\cite{li2021ai}. 
In our learning model, a dance motion is generated by sequentially producing 20 frames of dance with 120 frames of dance motion and 240 frames of onomatopoeic sequences. The sound symbolic word features and motion features are initially embedded in an 800-dimensional hidden representation within a linear layer. A learnable position encoding is then added before being input into the transformer layer.
This method allows for the generation of dance motions that correspond to the input onomatopoeic sequences, providing a unique and expressive way to create choreography. Despite potential limitations due to the size of the dataset or the AI Choreographer framework’s age, our implementation demonstrates the viability of using onomatopoeia as a medium for generating dance motions.


\section{Results and Evaluations}
\begin{table}[ht]
 \caption[Comparison of quantitative scores for each training model]{Comparison of quantitative scores for each training model. The larger the data set, the better the score when compared in terms of the size of the data set, and the music score was better when compared in terms of whether the feature was music or onomatopoeia.}
\label{table:SpeedOfLight}
 \centering
  \begin{tabular}{cllll}
   \hline
   Learned model &    FID\_k &    FID\_g &    Dist\_k &    Dist\_g\\
   \hline \hline
   AI Choreographer & 33.7593 & 15.4568 & 7.8050 & 6.4799 \\
   AI Choreographer Mini & 45.3925 & 26.4813 & 6.0897 & 5.5959 \\
   \hline
   our model & 81.8418 & 39.5038 & 2.0414 & 2.6655 \\
   \hline
  \end{tabular}
\end{table}


\subsection{Quantitative evaluation}
FID scores, which quantify the difference from the distribution of grand-truth motions, and motion diversity scores, which calculate the average Euclidean distance in the feature space across 40 generated motions on the AIST++ test set to measure the diversity were calculated for the three models: AI Choreographer, a previous study that generates dances from music; AI Choreographer Mini, which has the same training data as this model; and this model. Both the FID score and the diversity score were calculated for AI Choreographer and this model scored worse than AI Choreographer Mini. In other words, there is room for improvement in both the amount of data set and the way feature vectors are created.

\begin{figure*}[htbp] 
\centering
\includegraphics[width=0.8\linewidth]{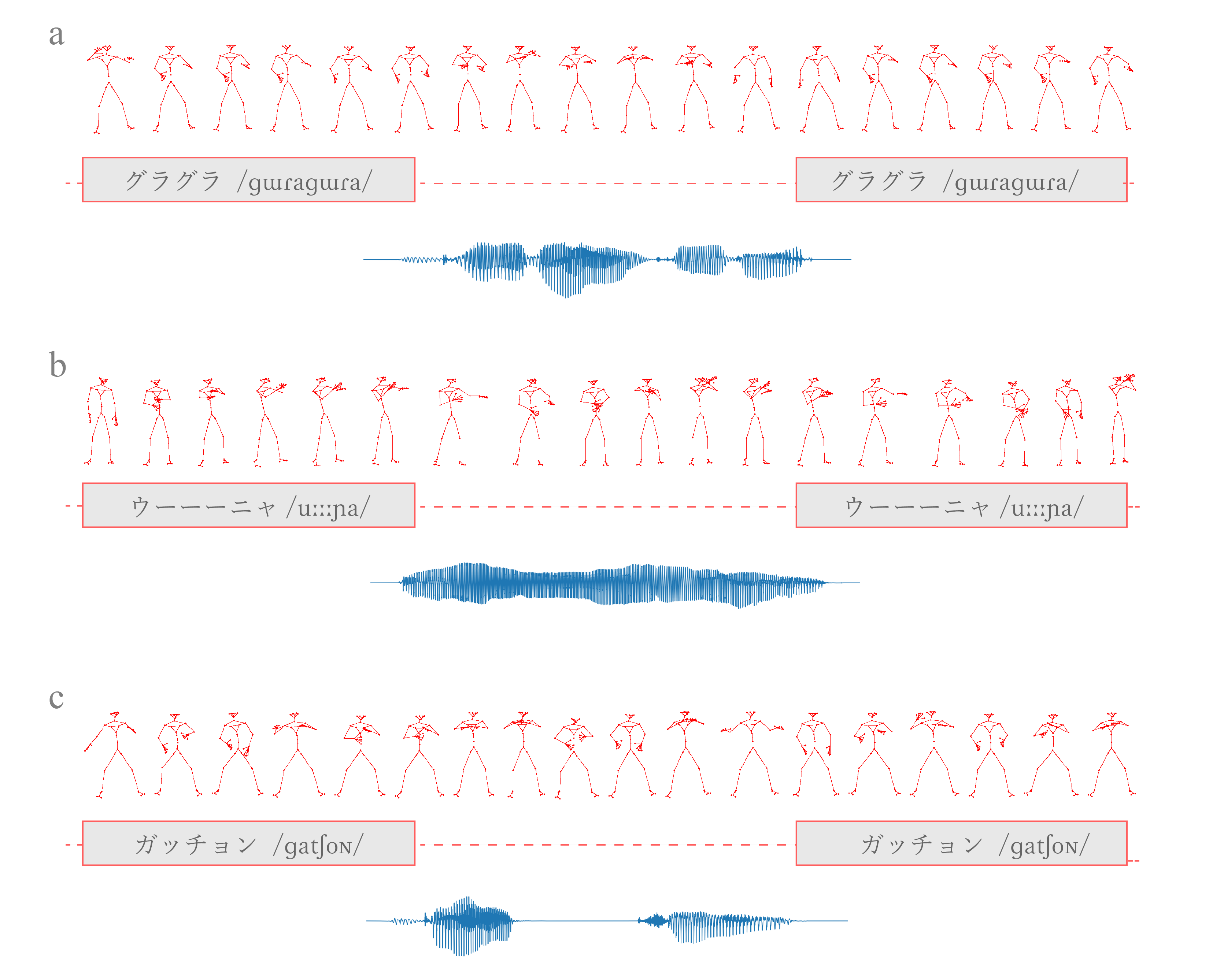}
\caption[results]{The dances generated with the onomatopoeia included in the validation data, and the waveform of the sound when that onomatopoeia occurred. The section where the onomatopoeia is described indicates the moment when the input embedding was given, and the dashed line represents the section where a zero vector was given. The frames where the dance and the characters of the onomatopoeia are written are time-aligned, but the waveform only shows the waveform of a single occurrence of the onomatopoeia.  The meaning of the sound symbolism of each onomatopoeia in ChatGPT is: a. The sound of some liquid boiling b. The sound of a cat meowing c. The sound of something hitting or being hit by a hard object.}
\label{fig:resultsval}
\end{figure*}




\subsection{Validity of the Generated Dance}

To verify the validity of the generated dance, we compare the input onomatopoeia's audio waveform with the generated dance. The audio waveform represents the rhythm of the onomatopoeia, and since the rhythm is closely related to the onomatopoeia meaning, we can investigate the semantic plausibility by looking at how the onomatopoeia meaning and waveform are related to the dance movement (see \autoref{fig:resultsval}). The first row (\autoref{fig:resultsval}a) represents the swaying motion in Japanese onomatopoeia, and from the waveform, we can see that there are four phonemes and a rhythm. Looking at the generated dance, we can see that the hands are swaying up and down with a slow, steady rhythm. The second row (\autoref{fig:resultsval}b) represents the sound of a cat stretching and meowing in Japanese onomatopoeia. From the waveform, we can see that the sound is being stretched. In the generated dance, the hands are stretched out and bent horizontally, reflecting the stretching image. The third row (\autoref{fig:resultsval}c) represents a Japanese onomatopoeia depicting the sound of heavy machinery with multiple parts hitting the floor or objects. From the waveform, we can see the bouncing sound. Looking at the generated motion, we can see the sharp lifting, lowering, and spreading of the hands, reflecting the bouncing sound. In this way, we could confirm that our method can generate reasonably plausible movements to a certain extent.

\subsection{Results in Various Languages}
To generate a dance intuitively even without knowledge of onomatopoeia, it is required that plausible dances can be generated even when onomatopoeia that comes to mind are used as is for input. The Sakamoto system we employed focuses on the phonemes and phonology of the input word, so there are virtually no restrictions on the onomatopoeia that can be used as input, and even created onomatopoeia can be used as input. To demonstrate this experimentally, we additionally generated a variety of onomatopoeia and attempted to generate dances.

\subsubsection{Onomatopoeia Generation in Various Languages}
We used the large language model, ChatGPT (GPT-4)~\cite{openai2023gpt4}, to generate onomatopoeia. ChatGPT can generate plausible text responses when given text-based instructions. We instructed it for the 26 languages it can handle to "Please generate onomatopoeia in XXX (language). However, for languages without onomatopoeia, please generate words that sound like onomatopoeia." We then had it generate four random onomatopoeia for each language.

\subsubsection{Results}
\begin{figure*}[htbp] 
\centering
\includegraphics[width=0.9\linewidth]{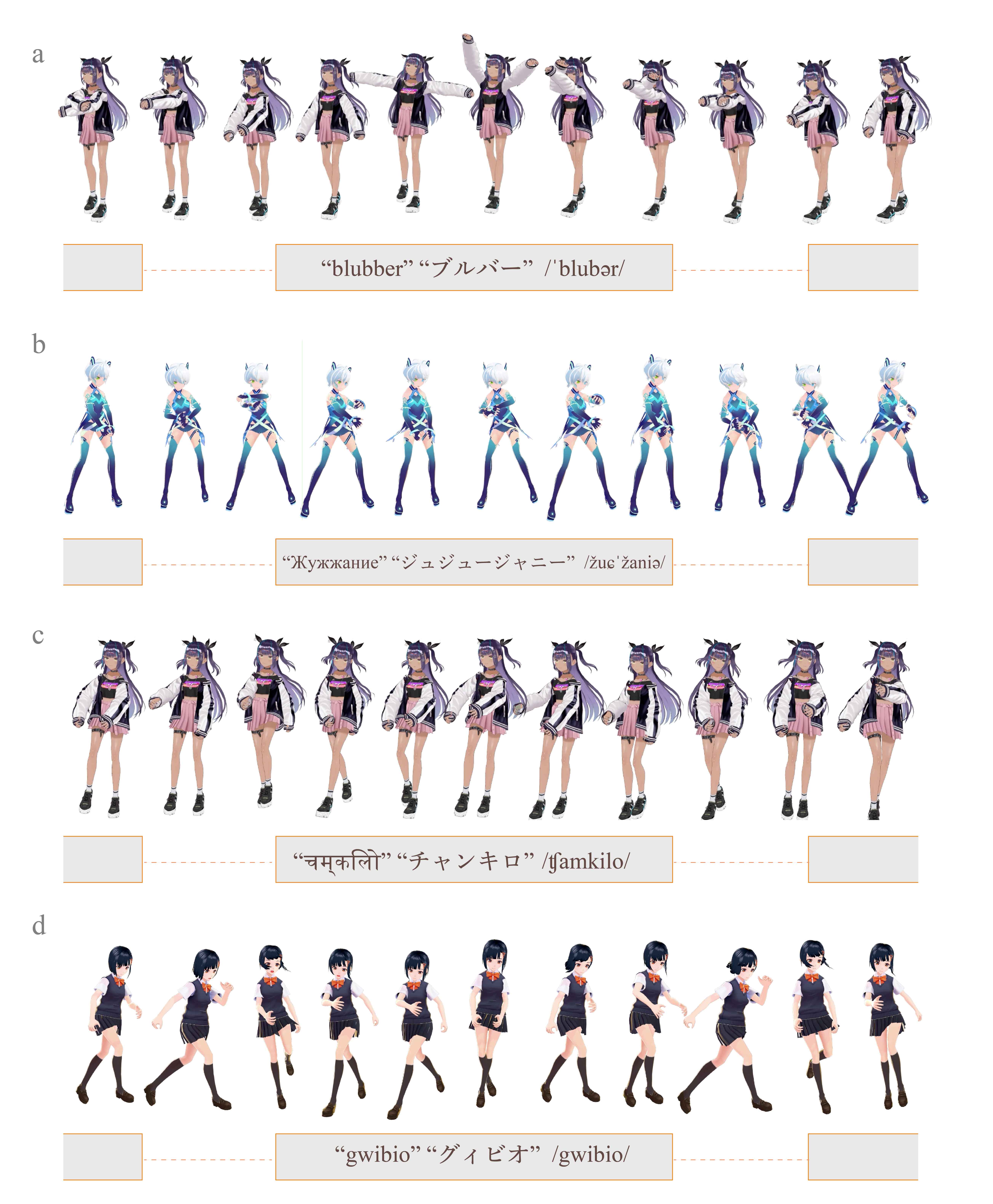}
\caption[results]{Generated motion results. The original language and the meaning of the sound symbolism of each onomatopoeia in ChatGPT is: a. German, Pukupukku and bubbling sound b. Russian, buzzing sound of insects c. Nepali, sound of shining light  d. Welsh, sound of wind blowing}
\label{fig:resultsgpt}
\end{figure*}

Examples of the generated dances that exhibited good movements are shown in \autoref{fig:resultsgpt}. For other examples, please refer to the supplementary materials. We confirmed that our model can generate plausible motions as dances when given unknown onomatopoeia as input. Additionally, we found that the model can generate similar motions repeatedly for repeated onomatopoeia inputs, indicating a consistency in the generated motions corresponding to the onomatopoeia. However, further verification is needed to determine if the appearance of the motions truly matches human impression.

\section{Discussion}
\subsection{Limitations and Future Work}

There are several limitations to this study that warrant further investigation. First, more updated model than AI Choreographer is available now which could generate more optimal dance motions. Future work should consider implementing more recent state-of-the-art methods to improve the quality of the generated motions. Additionally, the dataset used in this study is relatively small, which may have limited the performance of our model. In future studies, increasing the size of the dataset and fine-tuning the model can be considered to enhance the quality of the generated motion.

The proposed model has substantial potential applications in various domains. In the entertainment industry, this model can be used to create unique and engaging dance performances based on onomatopoeic input, adding a new dimension to the creative process. Furthermore, the model can be integrated into interactive applications, such as virtual reality or video games, to generate dynamic and immersive dance experiences for users.

In the field of education, this model can serve as a valuable tool for teaching dance and movement. By providing onomatopoeic input, students can better understand and visualize the relationship between rhythm, meaning, and movement in the context of dance. Moreover, this model can also be used to study the impact of different onomatopoeic words on dance motion generation and explore the underlying principles that govern this relationship.

\subsection{Uniqueness of onomatopoeia Approach}
We would like to discuss the following points: 1. The difference between generating onomatopoeia and random generation, 2. How the generation method differs from time-series data based on phonetic symbols, 3. The significance of using onomatopoeia as an intermediate representation between meaning, words, and sound. 
Firstly, the generation of onomatopoeia is different from random generation in that it carries both meaning and a sense of rhythm. Unlike random generation, onomatopoeia can convey specific information while still maintaining an abstract representation that allows for creative interpretation.
Secondly, the proposed method differs from time-series data generation based on phonetic symbols, as it does not solely rely on the intensity of sound. Instead, the model generates dance motions that consider both the rhythm and meaning of the input onomatopoeia. This makes the method more sophisticated than simply following the intensity of sound in a time-series manner.
Moreover, onomatopoeia serves as an intermediate representation between meaning, words, and sound. It is not strictly tied to the time-series data of sound intensity, allowing for more flexibility and expressiveness in the generated dance motions. This unique characteristic of onomatopoeia makes it a suitable input for generating dance movements that convey both rhythm and meaning.
Although the dataset presented in this study is still relatively primitive, we believe that it holds significant potential as an intermediate representation between meaning and physical information. By further refining the dataset and improving the model, it is possible to create more expressive and meaningful dance motions that fully utilize the unique properties of onomatopoeia.
Moreover the proposed method of using onomatopoeia for generating dance motions offers a novel and promising approach to computer graphics and animation research. By considering both rhythm and meaning, the model can generate more expressive and engaging dance performances that are not limited by the constraints of traditional time-series data or phonetic symbols. As the dataset and model continue to be developed, this research is expected to contribute to the advancement of the field and inspire new applications in various domains.

Another area for future work is the development of an interactive application that generates dance movements based on onomatopoeic input. Such an application would allow users to create personalized dance experiences and further explore the potential of onomatopoeia in the context of dance motion generation.

Qualitative evaluation will be conducted in the user study.
It is conceivable that the onomatopoeic instructions to be annotated could be squeezed or annotated with voice.


\section{conclusion}
In conclusion, this study has presented a novel approach to dance motion generation by leveraging the unique properties of onomatopoeia. The proposed machine learning model expands upon the existing AI Choreographer framework, introducing onomatopoeic input as a means to convey both rhythm and meaning in dance movement generation. The application of onomatopoeia as an input source is an innovative and linguistically interesting method, addressing the shortcomings of traditional dance generation inputs such as music, speech, and motion capture.

Despite its limitations, this study has demonstrated the potential of onomatopoeia as a medium for dance motion generation and has introduced a new dataset for further exploration. Future work could involve fine-tuning the model, increasing the dataset size, and developing an interactive application that generates dance movements in response to onomatopoeic input. This would enable users to easily create unique and expressive dance sequences based on their linguistic creativity.

\begin{acks}
This work was supported by READYFOR.
\end{acks}
\clearpage
\bibliographystyle{ACM-Reference-Format}
\bibliography{sample-base}



\end{document}